\documentclass{bvm} 

\usepackage{fancyhdr}

\begin{document}

\fancypagestyle{firstpage}{
  \fancyhf{}
  \fancyfoot[C]{Manuscript accepted at Bildverarbeitung für die Medizin (BVM) 2026}
  \renewcommand{\headrulewidth}{0pt}
  \renewcommand{\footrulewidth}{0pt}
}

\newcommand{\bvmyear}{2026}

\selectlanguage{english} 

\title{Kidney Cancer Detection Using 3D-Based Latent Diffusion Models}


\titlerunning{Kidney Cancer Detection Using Latent Diffusion}

\author{
    \fname{Jen} \lname[0009-0006-0632-7137]{Dusseljee},
     \fname{Sarah} \lname[0000-0001-5184-4340]{de Boer},
    \fname{Alessa} \lname[0000-0002-7602-803X]{Hering}
}


\authorrunning{Dusseljee, de Boer \& Hering}

\institute{
Department of Medical Imaging, Radboudumc, Nijmegen, The Netherlands
}

\email{sarah.deboer@radboudumc.nl}

\maketitle
\thispagestyle{firstpage}

\begin{abstract}

In this work, we present a novel latent diffusion-based pipeline for 3D kidney anomaly detection on contrast-enhanced abdominal CT. The method combines Denoising Diffusion Probabilistic Models (DDPMs), Denoising Diffusion Implicit Models (DDIMs), and Vector-Quantized Generative Adversarial Networks (VQ-GANs). Unlike prior slice-wise approaches, our method operates directly on an image volume and leverages weak supervision with only case-level pseudo-labels. We benchmark our approach against state-of-the-art supervised segmentation and detection models. This study demonstrates the feasibility and promise of 3D latent diffusion for weakly supervised anomaly detection. While the current results do not yet match supervised baselines, they reveal key directions for improving reconstruction fidelity and lesion localization. Our findings provide an important step toward annotation-efficient, generative modeling of complex abdominal anatomy.

\end{abstract}

\section{Introduction}
Kidney cancer has a yearly incidence rate of approximately 400,000 new cases worldwide. Since patients often remain asymptomatic until advanced stages, automatic detection of kidney tumors on computed tomography (CT) scans could positively affect patient outcomes. Artificial intelligence (AI) provides promising avenues in this field, and has been effectively applied to kidney lesion segmentation~\cite{3819-de2025robust}. However, most approaches rely on supervised learning requiring large annotated datasets that are time-consuming to create and may generalize poorly to rare abnormalities.

To address these problems, un- or weakly supervised methods can be used. These methods leverage the idea that lesions can be identified as deviations from healthy anatomy. Reconstruction-based methods use generative models to synthesize healthy image reconstructions. Subtracting the reconstruction from the input yields an anomaly map highlighting potential lesions. 
In particular, diffusion models, have recently demonstrated effectiveness in medical anomaly detection \cite{3819-wolleb2022diffusion,3819-wyatt2022anoddpm}.

Despite the prevalence of 3D architectures in medical imaging, most diffusion-based anomaly detection methods still process images slice by slice, potentially overlooking valuable inter-slice dependencies and volumetric structural information inherent in 3D imaging data.
To enable efficient 3D processing, latent diffusion \cite{3819-rombach_2022} enhances efficiency by operating in the compressed latent space of autoencoders, such as VQ-GANs \cite{3819-esser2021taming}, reducing computation while preserving image quality. Existing work has shown that, using latent diffusion, realistic 3D CT volumes can be generated, enabling the construction of fully volumetric generative pipelines \cite{3819-khader2023denoising}. Additionally, latent diffusion has shown promise for out-of-distribution detection~\cite{3819-graham2023unsupervised}, a task closely related to anomaly detection.

This work makes two main contributions: (1) a novel 3D weakly supervised anomaly detection framework combining DDIM, DDPM and VQ-GAN; and (2) a benchmark comparison with supervised methods for kidney abnormality detection, filling a gap in existing diffusion model literature.

\section{Materials and methods}
\subsection{Proposed method}
\begin{figure}[h]
    \centering
    \includegraphics[width=\linewidth]{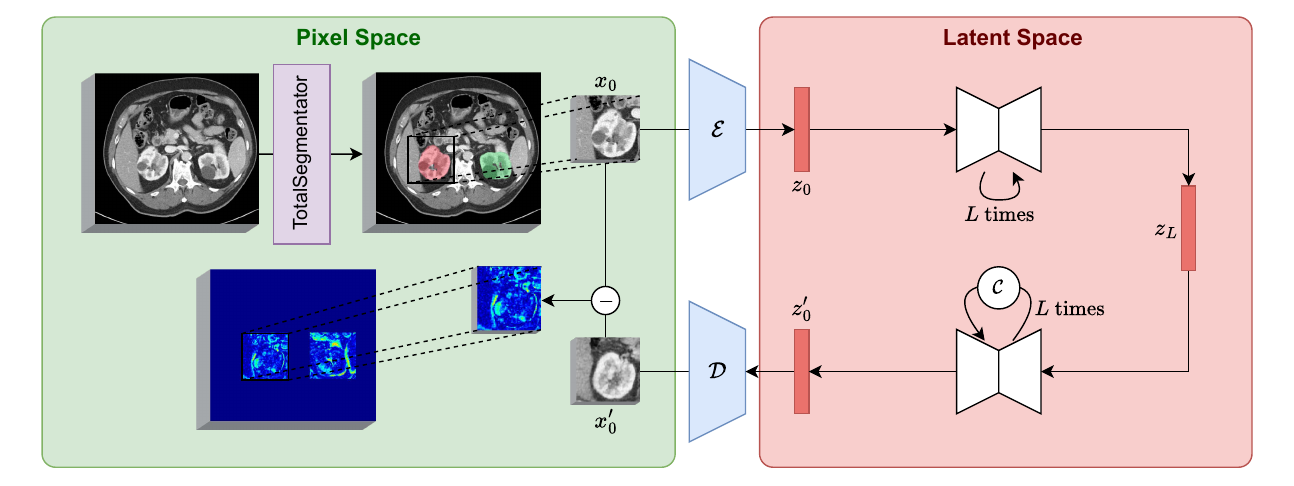}
    \caption{Overview of the proposed anomaly detection pipeline. Kidney patches, extracted using TotalSegmentator masks, are encoded into a latent space via a VQ-GAN. The latent patch undergoes $L$ forward (noising) diffusion steps, followed by $L$ reverse (denoising) steps with classifier guidance. The denoised latent patch is decoded to produce a healthy reconstruction. Subtracting this from the original yields the anomaly map.}
    \label{3819-fig:pipeline}
\end{figure}

We propose a novel method for kidney anomaly detection in full 3D CT images. Our pipeline combines diffusion-based anomaly detection~\cite{3819-wolleb2022diffusion} with an existing latent diffusion architecture~\cite{3819-khader2023denoising} for efficient 3D processing. An overview of the pipeline can be seen in \autoref{3819-fig:pipeline}. The following subsections describe each component in more detail, along with their design decisions.

\subsubsection{Pre-processing}
Each input CT scan is resampled to 1 mm isotropic resolution. To select the ROI, kidney segmentation masks are obtained using TotalSegmentator~\cite{3819-wasserthal2023totalsegmentator} (fast preset), and $96 \times 96 \times 128$ mm patches are extracted around each kidney. Focusing only on these areas leads to more efficient processing, while reducing false positives outside of the kidneys.

\subsubsection{Anomaly detection using latent diffusion models}
\paragraph{Diffusion-based anomaly detection}
We adopt Denoising Diffusion Probabilistic Models (DDPMs)~\cite{3819-ho2020denoising}, which generate realistic samples through an iterative denoising process. For anomaly detection, an input image is first corrupted with Gaussian noise via the forward diffusion process and subsequently reconstructed using the reverse denoising process of a model trained exclusively on healthy data. The resulting reconstruction represents a healthy approximation of the input. An anomaly map is then obtained by computing voxel-wise differences between the original image and the healthy reconstruction.

\paragraph{Implicit models for anatomically accurate reconstruction}
A key limitation of DDPM-based anomaly detection is the inherent stochasticity of the sampling process, which may introduce anatomical inconsistencies between the original image and its reconstruction. These deviations can lead to false positive detections and reduced anomaly detection performance~\cite{3819-wolleb2022diffusion,3819-wyatt2022anoddpm}. To address this limitation, we compare them with Denoising Diffusion Implicit Models (DDIMs)~\cite{3819-song2020denoising}. In contrast to DDPMs, the forward and reverse processes of DDIMs can be formulated as exact inverses, allowing for more anatomically accurate reconstructions~\cite{3819-wolleb2022diffusion}. However, due to the deterministic nature of DDIM sampling, classifier guidance~\cite{3819-dhariwal2021diffusion} is required to explicitly steer the reconstruction toward healthy tissue.

\paragraph{Latent diffusion for efficient 3D processing}
To enable efficient processing of 3D medical images, we apply latent diffusion~\cite{3819-rombach_2022}, where the diffusion process operates in a compressed latent space rather than directly in pixel space.
Specifically, we adopt the architecture introduced in \cite{3819-khader2023denoising}, which is designed for 3D medical image synthesis and employs a Vector-Quantized Generative Adversarial Network (VQ-GAN)~\cite{3819-esser2021taming} as the autoencoder model. Classifier guidance is performed directly within this latent space.

\subsubsection{Post-processing}
After performing anomaly detection for each kidney patch, we compose the results back into a full-scan anomaly map. This map is then resampled to the original resolution.
To reduce noise and false positives in the anomaly map, we apply morphological opening and closing, followed by a hole filling algorithm. Finally, we remove instances smaller than 20 voxels or <3mm in diameter. 

\subsection{Generating pseudo-labels}
The VQ-GAN, DDIM, DDPM, and classifier models are trained on a private dataset from Radboudumc. Contrast-enhanced thorax-abdomen/abdomen CT scans (slice thickness $\le 1$ mm) acquired between 2008 and 2021 were included, totaling 8,377 scans from 7,571 studies of 6,800 patients, yielding 5,095 left and 5,099 right kidneys. Pseudo-labels were derived from radiology reports: kidneys with reported lesions, cysts, or tumors were labeled unhealthy; those described as normal or unmentioned were labeled healthy. Kidneys with stones, calcifications, necrosis, atrophy, or prior removal were excluded. 

\subsection{Training details}
We implemented VQ-GAN and DDPM (with DDIM sampling and classifier guidance) based on~\cite{3819-khader2023denoising}. The VQ-GAN was trained for 100k iterations on the full dataset using an A100 GPU (batch size 4) in 1 day. The DDPM, with 1000 timesteps, was trained for 250k iterations on healthy kidney images (batch size 40) over 3 days on an A100. A U-Net–based classifier, using the encoder path with a linear head, was trained on a balanced pseudo-labeled dataset (1,162 healthy and 1,162 unhealthy kidneys) for 100 epochs (batch size 32) with early stopping, achieving a validation AUC of 0.70 and a test AUC of 0.56.

\subsection{Benchmarks and datasets}\label{3819-methods:datasets}
We compare our weakly supervised approach to two state-of-the-art supervised baselines. One model is used as-is from prior work, while the other is retrained to serve as a detection-specific benchmark.
The first baseline is a pretrained nnU-Net~\cite{3819-isensee2021nnu} for kidney and lesion segmentation \cite{3819-de2025robust}, trained with 5-fold cross-validation on the supervised dataset.
For detection benchmarking, we train nnDetection \cite{3819-baumgartner2021nndetection} with 5-fold cross-validation on the same data, reserving a $20\%$ subset as a holdout test set.
The baselines are trained on a combination of the KiTS23 \cite{3819-heller_2023} training set and the Radboudumc kidney abnormality dataset \cite{3819-humpire-mamani_2023}. Hyperparameter tuning for the proposed method is done on a validation fold from the nnDetection setup. 
A test set of 30 CT scans with fully annotated segmentation masks from Radboudumc was used as a test set to compare the proposed method, nnDetection and nnU-Net.

\section{Results}

\subsection{Detection and segmentation performance}
We measure segmentation performance using Dice similarity coefficient (DSC) and lesions are reported as detected when an intersection over union (IoU) of $0.2$ is reached.
A hyperparameter sweep for both DDPM and DDIM was conducted. The DDPM achieved the highest DSC on the validation set with a noise level of $L=500$ and classifier guidance strength $s=1600$. The DDIM achieved optimal performance at $L=500$ and $s=1800$. These optimal hyperparameters were subsequently used to evaluate segmentation performance on the hold out test set of the nnDetection dataset, and the private test set. The results are reported in~\autoref{3819-tab:segmentation_kits_radboudumc}. To determine the confidence threshold, we applied the Otsu method \cite{3819-otsuMethod}, as proposed in \cite{3819-wolleb2022diffusion}. 

\begin{table}[h]
    \centering
    \caption{Segmentation and Detection performance of optimal configurations for all methods on both test sets at an IoU threshold of $0.2$.}
    \label{3819-tab:segmentation_kits_radboudumc}
    
    \begin{tabular*}{\textwidth}{l@{\extracolsep\fill}llll}
    \hline
\textbf{nnDetection Test} & DSC $\uparrow$ &Precision $\uparrow$ & Recall $\uparrow$ & F1-score $\uparrow$ \\
\hline
    DDPM        & \textbf{0.12 ($\pm$0.10)} & 0.02 ($\pm$0.04) & 0.16 ($\pm$0.30) & 0.03 ($\pm$0.06) \\
    DDIM        & 0.07 ($\pm$0.09) & 0.01 ($\pm$0.09) & 0.04 ($\pm$0.18) & 0.02 ($\pm$0.10) \\
    nnDetection & N/A              & \textbf{0.51 ($\pm$0.33)} & \textbf{0.85 ($\pm$0.26)} & \textbf{0.63 ($\pm$0.26)} \\
\hline
    \textbf{Radboudumc Test} & DSC $\uparrow$ & Precision $\uparrow$ & Recall $\uparrow$ & F1-score $\uparrow$ \\
\hline
    DDPM        & 0.08 ($\pm$0.10) & 0.01 ($\pm$0.02) & 0.15 ($\pm$0.28) & 0.02 ($\pm$0.03)\\
    DDIM        & 0.08 ($\pm$0.11) & 0.02 ($\pm$0.10) & 0.03 ($\pm$0.07) & 0.02 ($\pm$0.05) \\
    nnDetection & N/A              & 0.51 ($\pm$0.28) & \textbf{0.73 ($\pm$0.28)} & 0.55 ($\pm$0.24) \\
    nnU-Net     & \textbf{0.68 ($\pm$0.25)} & \textbf{0.78 ($\pm$0.30)} & 0.67 ($\pm$0.29) & \textbf{0.69 ($\pm$0.26)} \\
\hline
    \end{tabular*}
\end{table}

\begin{figure}[h]
    \centering
    \includegraphics[width=\linewidth]{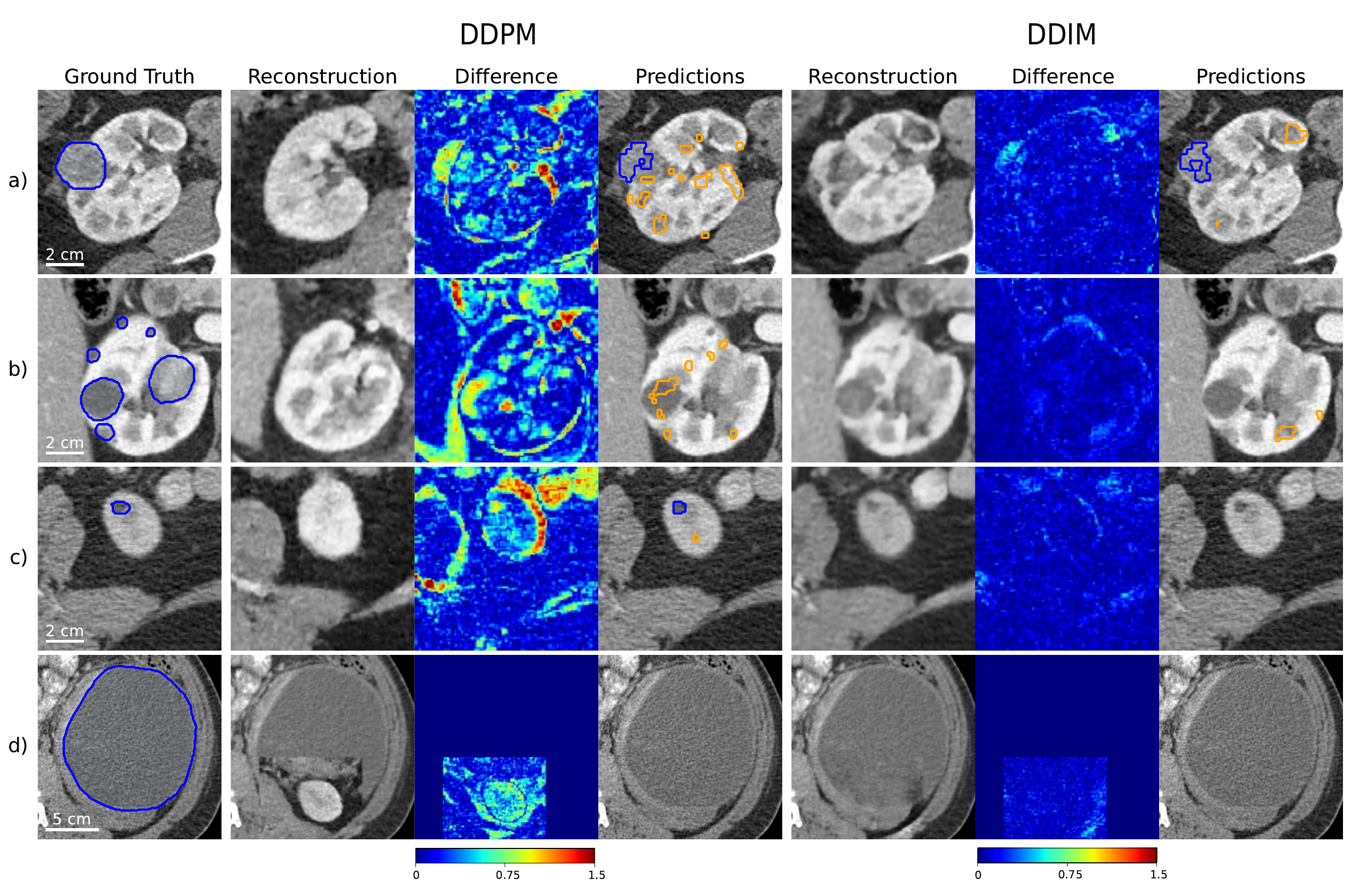}
    \caption{Visual comparison of diffusion-based reconstructions on four example images.}
    \label{3819-fig:visual-comparison}
\end{figure}

\subsection{Qualitative analysis}
\autoref{3819-fig:visual-comparison} presents a visual comparison between the reconstructions produced by the best-performing DDPM and DDIM models. The figure also shows the resulting difference map and the predictions after post-processing. Case a) shows a lesion present on the border of the kidney that was correctly detected by both models. Case b) shows multiple lesions within the kidney that were, none of which were detected by either model. Case c) shows a small lesion within the kidney that was found by our DDPM-based method, but missed by our DDIM-based method. Case d) shows a lesion that falls well outside the region of interest. Both models evidently struggle with this kind of lesion, with DDPM inpainting a completely new kidney.

\subsection{Evaluation at different lesion sizes}
\autoref{3819-tab:size_evaluation} shows the evaluation results at different lesion sizes. Evaluation was done using only the lesions in the reference masks that fall within the given size range. Remaining reference lesions with a matching prediction (IoU $\ge0.2$) were considered true positives, while the reference lesions without a matching prediction were considered as false negatives. Unmatched predictions within the size range were counted as false positives. Cases without reference lesions within the given size range were excluded. The number of remaining cases is reported within parentheses for each size range.

\begin{table}[t]
    \centering

    \caption{Segmentation and Detection performance of both unsupervised methods and nnDetection on the nnDetection test set for different lesion sizes at an IoU threshold of $0.2$.}
    \label{3819-tab:size_evaluation}

    \begin{tabular*}{\textwidth}{l@{\extracolsep\fill}lllll}
\hline
    Size (cm) & Model & DSC $\uparrow$ &Precision $\uparrow$ & Recall $\uparrow$ & F1-score $\uparrow$ \\
\hline
    \multirow{3}{*}{$\leq2$ (n $=69$)} &DDPM & \textbf{0.03 ($\pm$0.05)} & 0.01 ($\pm$0.02) & 0.14 ($\pm$0.29) & 0.02 ($\pm$0.04) \\
        & DDIM & 0.02 ($\pm$0.05) & 0.01 ($\pm$0.03) & 0.02 ($\pm$0.13) & 0.01 ($\pm$0.05) \\
        & nnDetection & N/A              & \textbf{0.18 ($\pm$0.24)} & \textbf{0.77 ($\pm$0.35)} & \textbf{0.44 ($\pm$0.24)} \\
    \hline
    \multirow{3}{*}{$2{-}4$ (n $=58$)} &DDPM & \textbf{0.09 ($\pm$0.14)} & 0.08 ($\pm$0.19) & 0.18 ($\pm$0.37) & 0.10 ($\pm$0.22) \\
        & DDIM & 0.03 ($\pm$0.11) & 0.02 ($\pm$0.1) & 0.05 ($\pm$0.22) & 0.03 ($\pm$0.14) \\
        & nnDetection & N/A              & \textbf{0.25 ($\pm$0.33)} & \textbf{0.94 ($\pm$0.22)} & \textbf{0.62 ($\pm$0.26)} \\
    \hline
    \multirow{3}{*}{$4{-}7$ (n $=37$)} &DDPM & \textbf{0.07 ($\pm$0.14)} & 0.09 ($\pm$0.28) & 0.11 ($\pm$0.31) & 0.09 ($\pm$0.28) \\
        & DDIM  & 0.00 ($\pm$0.02) & 0.00 ($\pm$0.00) & 0.00 ($\pm$0.00) & 0.00 ($\pm$0.0) \\
        & nnDetection & N/A              & \textbf{0.26 ($\pm$0.39)} & \textbf{0.84 ($\pm$0.37)} & \textbf{0.73 ($\pm$0.30)} \\
    \hline
    \multirow{3}{*}{$>7$ (n $=30$)} &DDPM & \textbf{0.02 ($\pm$0.06)} & 0.00 ($\pm$0.00) & 0.00 ($\pm$0.00) & 0.00 ($\pm$0.00) \\
        & DDIM & 0.00 ($\pm$0.02) & 0.00 ($\pm$0.00) & 0.00 ($\pm$0.00) & 0.00 ($\pm$0.00) \\
        & nnDetection & N/A              & \textbf{0.43 ($\pm$0.43)} & \textbf{0.72 ($\pm$0.44)} & \textbf{0.69 ($\pm$0.34)} \\
\hline
    \end{tabular*}
\end{table}

\section{Discussion}
While our diffusion-based methods did not yet match the performance of the supervised baselines in segmentation accuracy or detection sensitivity, this study provides valuable insights into the challenges and potential of diffusion-based reconstruction in complex anatomical regions. While prior work demonstrated viability in 2D brain MRI~\cite{3819-wolleb2022diffusion}, abdominal CT's greater variability in anatomy and contrast enhancement patterns may contribute to reduced performance. Both detection and segmentation performance are especially low for the largest lesion sizes, with DDIM failing to detect any lesions $\ge 4$ cm in diameter and DDPM failing for any lesion $\ge 7$ cm.

Visual inspection revealed that while our DDPM-based pipeline often correctly highlighted lesions, anatomical reconstruction artifacts frequently overshadowed true lesions, creating larger intensity differences than the lesions themselves. The false positives caused by these artifacts lead to low precision, especially when evaluating only lesions with a diameter $\le 2$ cm. Post-processing helped reduce false positives, but risked removing smaller lesions. Future work should investigate the use of false positive reduction networks~\cite{3819-hendrix2023deep}. Interestingly, DDIM with classifier guidance, which was introduced to mitigate anatomical reconstruction artifacts~\cite{3819-wolleb2022diffusion}, was outperformed by DDPM without classifier guidance, likely due to the limited performance of the classifier trained on pseudo-labels, which would have a stronger impact on DDIM's deterministic sampling than on DDPM's stochastic sampling. Simplex noise offers a promising alternative to Gaussian noise for reducing reconstruction artifacts in DDPM~\cite{3819-wyatt2022anoddpm}.

Additionally, alternatives to latent diffusion can be explored. Patch-based approaches~\cite{3819-pmlr-v227-bieder24a} operate directly in pixel space, eliminating the need for the classifier to operate in latent space, potentially leading to more effective guidance. Alternatively, wavelet diffusion~\cite{3819-phung2023wavelet} has been shown to be more efficient than latent diffusion, while leading to better results for healthy inpainting in brain MRI~\cite{3819-durrer2024denoising}.

We expect that performance can be significantly improved through a combination of these complementary strategies: false positive reduction, improved guidance mechanisms, and alternative diffusion architectures.




\begin{acknowledgement}
This research is funded by the European Union under HORIZON-
HLTH-2022: COMFORT (101079894). Views and opinions expressed are however those
of the author(s) only and do not necessarily reflect those of the European Union or European Health and Digital Executive Agency (HADEA). Neither the European Union
nor the granting authority can be held responsible for them.	
\end{acknowledgement}

\printbibliography

\end{document}